\RequirePackage{fix-cm}
\documentclass[a4paper]{svjour3}                     
\usepackage{times}
\usepackage{graphicx}
\usepackage{amsmath}
\usepackage{amssymb}
\usepackage{multirow}
\usepackage{subcaption}
\usepackage{color}
\usepackage{url}
\usepackage{fixltx2e}
\usepackage{xspace}

\newcommand{\van}{VAN\xspace}                     
\newcommand{\vanp}{VAN\textsubscript{p}\xspace}   
\newcommand{\vani}{VAN\textsubscript{i}\xspace}   
\newcommand{\vano}{VAN\textsubscript{o}\xspace}   
\newcommand{\ie}{{\it i.e.}}
\newcommand{\etal}{{\it et al.}}

\begin{document}

\title{Temporal Action Localization with Variance-Aware Networks
}


\author{Ting-Ting Xie         \and
        Christos Tzelepis     \and 
        Ioannis Patras 
}


\institute{Authors are with the Department
of School of Electronic Engineering and Computer Science, Queen Mary University of London, London, UK. \\
\email{t.xie@qmul.ac.uk, c.tzelepis@qmul.ac.uk, i.patras@qmul.ac.uk.}}

\date{Received: date / Accepted: date}

\maketitle

\begin{abstract}

This work addresses the problem of temporal action localization with Variance-Aware Networks (\van), \ie, DNNs that use second-order statistics in the input and/or the output of regression tasks. We first propose a network (\vanp) that when presented with the second-order statistics of the input, \ie, each sample has a mean and a variance, it propagates the mean and the variance throughout the network to deliver outputs with second order statistics. In this framework, both the input and the output could be interpreted as Gaussians. To do so, we derive differentiable analytic solutions, or reasonable approximations, to propagate across commonly used NN layers. To train the network, we define a differentiable loss based on the KL-divergence between the predicted Gaussian and a Gaussian around the ground truth action borders, and use standard back-propagation.
Importantly, the variances propagation in \vanp does not require any additional parameters, and during testing, does not require any additional computations either. In action localization, the means and the variances of the input are computed at pooling operations, that are typically used to bring arbitrarily long videos to a vector with fixed dimensions. 
Second, we propose two alternative formulations that augment the first (respectively, the last) layer of a regression network with additional parameters so as to take in the input (respectively, predict in the output) both means and variances. Results in the action localization problem show that the incorporation of second order statistics improves over the baseline network, and that \vanp surpasses the accuracy of virtually all other two-stage networks without involving any additional parameters.\footnote{We will release the code upon acceptance of the manuscript.}
\end{abstract}


\section{Introduction}
\label{intro}
In the recent years there has been a tremendous interest in video analysis tasks such as action recognition~\cite{simonyan2014two,tran2015learning,wang2016temporal,carreira2017quo,feichtenhofer2018slowfast} and action localization~\cite{gao2017cascaded,chao2018rethinking,lin2019bmn,xu2020g}. The latter is concerned with determining not only which actions are depicted in a video, but also finding their temporal borders, that is, their start and end. This is particular important in long, untrimmed videos that are characteristic of user generated and not curated data. 

\begin{figure}[t]
\centering
    \includegraphics[width=0.65\linewidth]{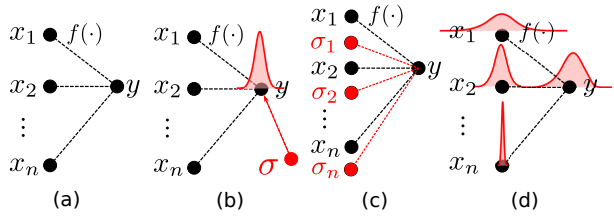}
    \caption{Illustration of the proposed methods versus baseline for the regression task of the form $y=f(\mathbf{x})$, where $\mathbf{x}=(x_1,x_2,\ldots,x_n)^\top$ are the input features. (a) Baseline network uses features $\mathbf{x}$ and predicts $y$. (b) \vano uses the same features and predicts $y$, which is then modeled as a uni-variate Gaussian $\mathcal{N}\left(y,\sigma^2\right)$ with $\sigma$ being introduced as an additional learnable parameter. (c) \vani exploits the aforementioned input variances as additional features and uses them similarly to the baseline. (d) \vanp, the input features are modeled as uni-variate Gaussians, $\mathcal{N}\left(x_i,\sigma_i^2\right)$, with $\sigma_i$'s being calculated by the pooling operation. Input variances and means are then \emph{propagated} throughout the network so as to deliver probabilistic predictions in the output in the form of the mean and the variance. It is worth noting that \vani requires almost as many as double the number of parameters of baseline, while \vanp requires no extra parameters.}
    \label{fig:proposed_method_intro}
\end{figure}

Several works~\cite{gao2017turn,gao2017cascaded,chao2018rethinking,lin2019bmn} in action localization follow a two-stage approach, in which a proposal generation network generates sparse, class-agnostic proposal candidates in the first stage, and a classification network classifies those candidates into one of the several action classes at the second stage. In virtually all of those works, the second-stage brings video segment of variable length into a fixed dimension by a pooling operation, such as a standard max/average pooling over regular bins, as in ~\cite{shou2016temporal,shou2017cdc,gao2017cascaded,gao2017turn,xie2019exploring} or bins defined over a hierarchical structure, as in Structured Temporal Pyramid Pooling (STPP)~\cite{zhao2017temporal}. While pooling has proven to be extremely effective in reducing input dimensionality and providing feature maps robust to small spatial/temporal transformation, it also results to loss of information as it summarizes the features within the pooling area by their mean (or maximum) value.

In this paper, we first propose a Variance-Aware Network, which we call \vanp, that utilizes, not only the standard first-order moments, \ie, the mean values, but also the second-order moments, \ie, the corresponding variances of the input, computed by standard average pooling. In the proposed method, we derive the means and the variances of the outputs of commonly used layers of DNNs, such as linear layer and ReLU, as function of the means and the variances of the inputs. These derivations can be either analytical or approximations. In this way, \vanp can propagate in a forward pass, the means and variances of the input layer all the way through the DNNs until the output of the last layer where we obtain means and variances for the prediction of the location of the temporal borders of the action. There, we define an appropriate differentiable loss between a Gaussian represented by the predicted mean and variance, and a Gaussian with a small variance that is defined around the ground truth prediction, namely the KL-divergence between the two Gaussians. Since all operations are differentiable, the error can be back-propagated and the network (\vanp) can be trained in an end-to-end fashion. The propagation does not introduce additional trainable parameters and does not require additional computational cost at test time. 

Moreover, we propose two additional formulations, \vani and \vano, that take into consideration variances either in the input or, respectively, in the output, by augmenting with additional parameters the corresponding layers. Inspired by~\cite{kendall2017whatuncertainties,he2019bounding}, \vano extends the latter~\cite{he2019bounding} to the problem of action localization. Clearly, \vani and \vano require more trainable parameters the number of which depends on the dimensionality of the input and output of the first (respectively last) layer of the baseline network. In our case, \vani requires almost the double number of parameters compared to the baseline, while \vano requires only a few more.


We show that in the problem of action localization, all Variance Aware Networks, consistently improve the baseline second stage network, and that \vanp typically performs better than the two variants that consider variances only in the input or output. In Fig.~\ref{fig:proposed_method_intro} we give an illustration of the proposed methods versus the baseline. 

The contributions of this paper can be summarized as follows:
\begin{itemize}
    \item We propose three Variance-Aware probabilistic prediction models for the problem of action localization; we are the first to propose to utilize the variance of the features that is typically lost during pooling operations in the problem of action localization.
    
    \item We show that, without additional parameters, second-order moments of the input of a DNN can be propagated all the way up to the output layer, and, once an appropriate loss function is defined, back-propagated so as to train it in an end-to-end fashion. To the best of our knowledge, this is the first work that does that.
    
    \item Comprehensive experiments show that the proposed \van can improve the performance of temporal action localization and enhance the robustness of the network. 
\end{itemize}  

\section{Related Work}\label{sec:rel_work}

In this section we will first review related work in the domain of action localization, focusing on state-of-the-art two-stage approaches. We, then, review works that are broadly related to our contribution in modeling and propagating second-order moments in Deep Neural Networks.

\subsection{Temporal Action Localization}

Action localization has greatly benefited from the progress in the domain of action recognition \cite{simonyan2014two,tran2015learning,wang2016temporal,carreira2017quo}, in particular due to the development of large datasets~\cite{UCF101,caba2015activitynet,kuehne2011hmdb,kay2017kinetics,gu2017ava}, which are used to train DNNs that serve as backbones or feature extractors for the problem of action localization. Such features include appearance and motion features extracted by two-stream networks~\cite{simonyan2014two,wang2016temporal,carreira2017quo,feichtenhofer2018slowfast} or spatio-temporal descriptors, such as C3D~\cite{tran2015learning}, that are extracted from deep 3-dimensional convolutional networks (3D ConvNets).

Temporal action localization methods both classify the action and detect its temporal boundaries in untrimmed videos. Some works, such as~\cite{ma2016learning,singh2016multi} use Recurrent Neural Networks to model temporal dependencies, however, their performance in comparison to CNN-based methods, possibly due to vanishing gradient in training, remains low.

Inspired by the great success in object detection~\cite{girshick2015fast,ren2015faster,he2017mask,cai2018cascade,he2019bounding}, most temporal action localization methods adopt a two-stage approach in which, at the first stage, class-agnostic temporal proposals of variable length are generated, and in the second stage, those proposals are assigned a class label and their borders are refined. While some works focus on generating better proposals~\cite{buch2017sst,gao2017turn,lin2018bsn,gao2018ctap,xu2019two,lin2019bmn}, other works~\cite{zhao2017temporal,gao2017cascaded,chao2018rethinking,girdhar2018video} focus on the classification/regression second stage. 

Typically, the latter uses as input a fixed-size input feature that is extracted from the variable length proposals by pooling~\cite{shou2016temporal,shou2017cdc,gao2017turn,gao2017cascaded}. However, global pooling discards information and a few methods have been proposed to partially address this issue. In~\cite{zhao2017temporal,xie2019exploring}, the authors either use Structured Temporal Pyramid Pooling (STPP) or Part-divided temporal pooling, that is, they perform pooling at different locations and temporal durations -- however, while the temporal structure is better preserved, a classical average pooling is performed. Lin \etal~\cite{lin2018bsn} constructs the fixed-size feature by linear interpolation with 16 points -- this disregards completely the information at all but 16 temporal locations. Finally, Chao \etal~\cite{chao2018rethinking} performs filtering at the original resolution with a multi-tower network by using dilated filters of different dilation steps so as to better preserve the temporal structure. 

Finally, there exist approaches that optimize the aforementioned two stages jointly, such as one-stage detectors~\cite{lin2017single,buch2017end,huang2019decoupling}, REINFORCE-based methods~\cite{yeung2016end}, Graph neural network-based approaches~\cite{xu2020g,zeng2019graph}, or using Gaussian kernels to dynamically optimize the temporal scale of action proposals~\cite{long2019gaussian}. In this work we focus on two-stage methods.

\subsection{Uncertainty estimation}

Recently, considerable research effort has been directed towards measuring and utilizing second-order moments or statistics in the area of deep learning. Gal and Ghahramani~\cite{gal2016dropout} model the uncertainty of the weights of the networks using second-order statistics and utilizes it to build a probabilistic interpretation of dropout -- by contrast, we model second-order moments of the input and utilize them as additional features or propagate them across the network. Goroshin \etal~\cite{goroshin2015learning}, address the problem of inherent uncertainty in prediction by introducing into the network architecture latent variables that are non-deterministic functions of the input. In~\cite{girdhar2017attentional}, the authors build on the work of~\cite{carreira2012semantic} in order to develop an attention pooling mechanism that exploits second-order statistics for the problem of action recognition. In~\cite{kendall2017whatuncertainties}, authors study the benefits of modeling uncertainty in Bayesian deep learning models for vision tasks. In~\cite{shi2019pfe} the authors represent each face image as a Gaussian distribution in the latent space. In~\cite{he2019bounding}, they propose to use a KL-based loss as bounding box regression loss for learning bounding box transformation and localization variance jointly, and in~\cite{he2019deep} they propose a deep multivariate mixture of Gaussians model for probabilistic object detection under occlusion.

However, none of the above works compute second-order statistics analytically -- on the contrary, they use low-rank approximations so as to generate latent feature space. Abdelaziz \etal~\cite{abdelaziz2015uncertainty} propose a method for approximately propagating input uncertainty (residual noise and distortion) through a DNN using Monte Carlo sampling, for the problem of Automatic Speech Recognition (ASR). This work relies on sampling (rather than analytical derivations) for uncertainty propagation and learning, which can be prohibitively expensive in the case of high-dimensional input, which is typically the case in temporal action localization. Our work is also related to a recent work by Tzelepis \etal~\cite{tzelepis2018linear} that models input uncertainty using second-order moments. However, this method utilize input variance in a shallow, SVM-based max-margin framework with a loss that is the expectation of the hinge loss for classification -- by contrast, we address the problem of action localization, we introduce a new regression loss that is based on the KL-divergence and utilize the variances in a Deep Neural Network scheme that allows end-to-end training.

Finally, while Kullback-Leibler (KL) divergence has been extensively used in a variety of Machine Learning problems, including  matrix factorization~\cite{cichocki2006csiszar}, domain adaptation, weight regularization in DNNs, and optimization tasks~\cite{basseville2013divergence}, to the best of our knowledge it has never been used in the problem of temporal action localization.

\section{Variance Aware Network (\van)}\label{sec:proposed_method}

\begin{figure*}[t]
    \begin{center}
    \includegraphics[width=\linewidth]{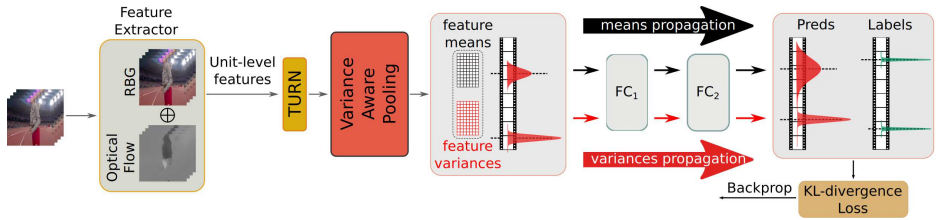}
    \end{center}
    \caption{Overview of Variance-Aware Network (\vanp): Given an untrimmed video, unit-level features are extracted. After that, detector takes proposals generated by a TURN~\cite{gao2017turn} as input, and outputs classification scores and and two regression offsets (start/end); this is depicted in the black solid stream (baseline). Variances are computed by Variance-Aware Pooling (in the cases of \vani and \vanp), or introduced as extra trainable parameters at the output (in the case of \vano). In the former case, variances are (i) either propagated across the network, up to its output, obtaining predictions as pairs of means and variances -- this is depicted in the red solid stream (\vanp), or (ii) concatenated to the means and used as features for training in a standard fashion at the input (after Variance-Aware Pooling operation). During testing, the predicted temporal boundaries are adjusted in a cascaded way by feeding the refined clips back to the system for further boundary refinement. All the parameters in each cascade step are shared.}
    \label{fig:overview}
\end{figure*}

In this section we describe our Variance Aware Networks (VAN) for action localization. We first briefly discuss the baseline, two-stage approach that we adopt, at the heart of which is a classification/regression network with pooling, linear and non-linear layers. Next, we present our proposed method for calculating both means and variances at the pooling layer and introduce \vani, where variances are used in the input using again extra, trainable parameters in the first layer and \vano, where variances at output are learnt using extra trainable parameters at the last layer. Next, we present our proposed method for propagating both means and variances throughout the network until the output, where we define an appropriate regression loss that allows for end-to-end training (\vanp). The outline of the proposed method is shown in Fig.~\ref{fig:overview}.

\subsection{Baseline method}\label{subsec:baseline_method}

The proposed Variance-Aware Network (VAN) builds on a two-stage approach baseline. The first stage network is a proposal generation network that takes as input video segments that have been produced by a sliding window approach and from which unit-level features are extracted, as in~\cite{wang2017untrimmednets}. This stage performs i) a binary (class-agnostic) classification task (on whether a segment depicts an action or not) and assigns a classification score to each input segment, and ii) a regression task for adjusting the borders of input segments. Based on the classification score, a top-ranked list of such segments are selected as action proposals feed to second stage network to perform temporal action localization. Along with these proposals, we also use a number of units, before and after the actual proposals, in order to capture context information. The latter has proven to be very useful in the action boundary detection task~\cite{gao2017cascaded,chao2018rethinking}.

The second stage network takes as input action proposals generated at first stage as discussed above. At this stage we perform i) multi-class classification of input proposals, so as to decide on the action class that the proposals belong to and ii) regression for adjusting the temporal borders of input proposals. For doing so, we first apply a pooling operation in order to fix the dimension of the input (proposals vary a lot in their temporal length/number of units). Pooling is typically used to this end, \ie, fixing the dimensionality of input features, but this comes at some cost. More specifically, a pooling operation discards any temporal information of its input, while average pooling may also mask salient parts of the video. To partially preserve the temporal structure, we divide each input proposal into $k$ parts as in~\cite{xie2019exploring}. Then we perform average pooling to each part and concatenate the resulting features. Additionally, we perform global pooling to the context parts and concatenate all features together. This results in a fixed-dimensional feature representation scheme for all input proposals. Finally, we use a sub-network comprised of two fully-connected layers (along with a $\mathcal{L}_2$-normalization layer and a ReLU) in order to obtain predictions for the classification and the localization part. In what follows, we focus on the second stage network - the structure of the baseline network is depicted in Fig.~\ref{fig:overview}, when the building blocks are classical DNN layers.

\subsection{Variance Aware Networks with layer augmentation at input (\vani) or output (\vano)}

In this subsection we will introduce two Variance Aware Networks that take into consideration second-order statistics either in the input (\vani), or in the output (\vano). Let us denote with $x\in\mathbb{R}^{d_i}$, and $y\in\mathbb{R}^{d_o}$ the input and, respectively, the output of the corresponding baseline network.

\subsubsection{\vani}

\begin{figure}[t]
\centering
    \includegraphics[width=0.65\linewidth]{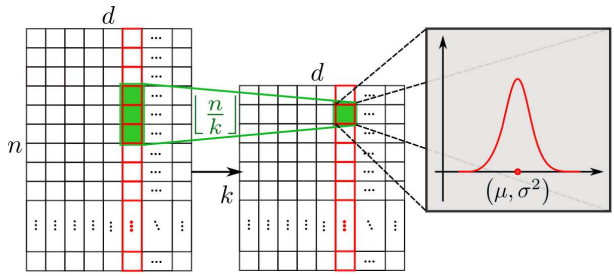}
    \caption{Variance-Aware Pooling (VAP) layer: first- and second-order moments are computed during the pooling operation. Output of pooling is given as pairs of means and variances.}
    \label{fig:variance_computation}
\end{figure}

As discussed in Sect.~\ref{subsec:baseline_method}, a pooling layer is typically used to bring a video of arbitrary length (\ie, arbitrary number of video units) to a feature vector $x\in\mathbb{R}^{d_i}$ of fixed number of dimensions $d_i$. We note the variation is nearly two orders of magnitude, since the proposals' length is ranging from dozens to thousands. The pooling operation naturally leads to loss of information, since all values within a bin are reduced to a single one. We partially compensate for this loss of information by computing not only the mean values (similarly to a standard average pooling operation), but also the corresponding variances. We refer to this, as a Variance-Aware Pooling (VAP) layer and we illustrate it in Fig.~\ref{fig:variance_computation}.

After a VAP layer, features are in the form of pairs of means and variances $(\mu_x,\sigma_x)$. \vani, the first of our proposed Variance Aware Networks, straightforwardly uses the variances as extra features in the inputs (\ie, by concatenating them with the means) and is trained in a standard manner, similarly to Sect.~\ref{subsec:baseline_method}. \vani uses a richer feature representation, which comes at the cost of doubling the parameters in the first FC layer -- in the specific baseline network that we use, this leads to almost doubling the parameters of the whole network. At test time, \vani receives pairs of values $(\mu_x, \sigma_x)$ in the input, where $\mu_x\in\mathbb{R}^{d_i}$ and $\sigma_x\in\mathbb{R}^{d_i}_+$, and delivers predictions $y\in\mathbb{R}^{d_o}$.

\subsubsection{\vano}

The second Variance Aware Network that we propose, namely \vano, is an extension of~\cite{he2019bounding} to the problem of action localization, and considers variances in the output. More specifically, \vano introduces an additional head in the output, and delivers a pair of predictions $(\mu_y,\sigma_y)$ that can be interpreted as means and variances of Gaussian distributions. That is, each prediction $y$ is defined as a uni-variate Gaussian $y\sim\mathcal{N}(\mu_y,\sigma_y)$. It is worth noting that, similarly to~\cite{he2019bounding}, here we consider only uni-variate Gaussians and, therefore, $\mu_y\in\mathbb{R}^{d_o}$ and $\sigma_y\in\mathbb{R}^{d_o}_+$, where $d_o$ is the number of dimensions of the output of the corresponding baseline network. In the case of action localization, $d_o=2$ as, for each proposal, we predict the start and the end of the action. In this framework, a predicted variance could be interpreted as the uncertainty of the corresponding predicted mean. In order to train the corresponding network, an appropriate differentiable loss function is defined. In this paper, we use the KL-divergence between the Gaussian at the output of the network, and a Gaussian with a small variance that is defined around the ground truth annotations (start and end of the actions). Then, the network can be trained with standard back-propagation. At test time, given an input vector $x\in\mathbb{R}^{d_i}$, \vano delivers a prediction $(\mu_y,\sigma_y)$ -- the variance $\sigma_y$ may, or may not, be used (in this paper, we don't).

\subsection{Variance propagation from input to output (\vanp)}\label{subsec:var_estim_and_prop}

In this section we will describe a network that propagates second-order statistics throughout the network, \ie, from the input to the output layer. As in \vani, we model the output of pooling as a set of uni-variate Gaussian distributions, for which we know their means and variances (and thus we have defined them uniquely). Then, we propagate these distributions (in terms of their first- and second-order moments) through the various network components and come up with predictions at its output that are also in the form of means and variances pairs and are also Gaussians. We refer to this network as \vanp, and in contrast to \vani and \vano, it requires no extra trainable parameters. To propagate through the network up to the last layer, we modify a number of typical DNN building blocks, like FC, ReLU, and normalization layers, as described below.
\paragraph{\textbf{Linear layers}} In the case of a linear layer, such as a FC layer, parametrized by a weights matrix $W$ and a bias terms vector $b$, the output means $M_{out}$ and variances $V_{out}$ are derived analytically with respect to input means $M_{in}$ and variances $V_{in}$ as follows:
\begin{align}\label{eq:fc_prop}
    M_{out} = & W^\top M_{in} + b \\
    V_{out} = & W^\top V_{in}W.
\end{align}
Irrespectively of whether the input covariance matrix ($V_{in}$) is diagonal or a full one, the output covariance matrix ($V_{out}$) will be a full one, since a linear operation like $W^\top X + b$ correlates input variables. This means that after propagation of variances through the first FC layer, we may have and propagate full covariance matrices. However, the memory requirements make this prohibitive for anything but toy networks as the size of the covariances are quadratic with the number of features. For this reason, in our work, we assume and propagate only diagonal covariance matrices; that is, after a FC layer, we keep only the diagonal part of the propagated variance matrix; \ie, $V_{out}$ in (\ref{eq:fc_prop}). This can be efficiently calculated as follows: 
\begin{equation}
    \operatorname{diag}(V_{out})= {W^2}^\top\operatorname{diag}(V_{in}),
\end{equation}
where $W^2$ is the element-wise square of the weights matrix $W$, and $\operatorname{diag}(\cdot)$ is an operator that acts on matrices and returns their diagonal parts as vectors. Clearly, there is no need to maintain or store the full covariance matrices.

\begin{figure}[t]
\centering
    \includegraphics[width=0.65\linewidth]{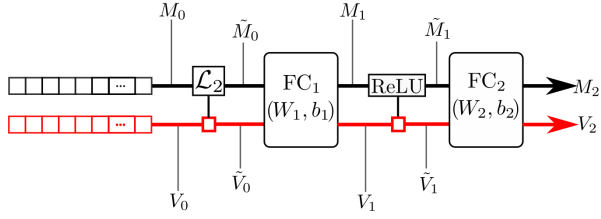}
    \caption{First-order (in black) and second-order (in red) moments propagation through the network. Output of the network (classification scores and regression values) are given as pairs of means and variances.}
    \label{fig:variance_propagation}
\end{figure}

\paragraph{\textbf{Non-linear layers}} Computation of a non-linear transformation of a Gaussian random variable in closed-form is often intractable. Even conceptually simple functions of a random variable are hard to be derived analytically or be efficiently computed. For this reason, for the non-linearities that we use in our network, we adopt the following approximations and/or simplifications:
\begin{itemize}
    \item \textbf{$\mathcal{L}_2$-normalization}: We first apply $\mathcal{L}_2$-normalization to the mean vectors as computed by the pooling layer ($M_0$ in Fig.~\ref{fig:variance_propagation}) and keep the (squared) norms of each mean vector in order to scale the corresponding variance vectors (by dividing each vector by the corresponding square norm).
    
    \item \textbf{Rectified Linear Unit (ReLU)}: The propagation via a ReLU layer ($y=\max(0,x)$), leads to means and variances at the output that can be expressed as follows:
    \begin{align}
        \mu_y      & =\int_{\mathbb{R}}\max(0,x)f(x)\mathrm{d}x    \\
        \sigma_y^2 & =\int_{\mathbb{R}}\max(0,x)^2f(x)\mathrm{d}x.
    \end{align}
    where $f$ denotes the PDF of $x\sim\mathcal{N}\left(\mu_x,\sigma_x^2\right)$. While analytical expressions can be derived, they are quite complex. For this reason, we decided to apply a standard ReLU operation on the means and keep only the variances for which means are non-negative ($\tilde{M}_0$ and $\tilde{V}_0$ in Fig.~\ref{fig:variance_propagation}).
\end{itemize}

The modified network of \vanp is shown in Fig.~\ref{fig:variance_propagation}. The upper branch of the network (depicted in black) is responsible for propagating the means (and is essentially the same as the baseline network described in Sect.~\ref{subsec:baseline_method}) and the lower branch (depicted in red) is responsible for propagating the variances. These moments are propagated through a number of linear (FC) and non-linear ($\mathcal{L}_2$-normalization, ReLU) layers before arriving at the output. Crucially, all the computations at the lower branch do not require additional trainable parameters in the network, but do require additional activation maps so as to store the variances. 

As we will show in the experimental section, both the mechanisms, \ie, concatenation of the means and variances in the input on the one hand (\vani), and propagation of the means and variances on the other (\vanp) result in consistent and large improvements over the baseline architecture. 



\subsection{Kullback-Leibler divergence as an uncertainty-aware regression loss}\label{subsec:kl_loss}

Two variants of the proposed method, \ie, \vanp and \vano, model output localization predictions as uni-variate Gaussian distributions; that is, as pairs of means and variances (see Fig.~\ref{fig:proposed_method_intro} and Fig.~\ref{fig:variance_propagation}). In the case of regression to the action temporal boundaries, these are means and variances of our predictions $p$ about the start and the end of the action. The prediction $p$ can then be naturally thought to follow Gaussian distributions with those means and variances, that is $p\sim\mathcal{N}\left(\mu_p,\sigma_p^2\right)$. In order to define an appropriate loss, we consider that the ground truth action borders also follow Gaussian distributions with known means, as given by the human annotators, and variances that we artificially set to a small value. The variances  could be thought as expressing the degree of uncertainty that is introduced by the annotation process. Let then  $t\sim\mathcal{N}\left(\mu_t,\sigma_t^2\right)$ be the empirical distribution of the ground truth values, where $\mu_t$ are the ground truth annotations and $\sigma_t^2$ is set to 0.01.

A natural measure of dissimilarity between two distribution is the Kullback-Leibler (KL) divergence, that in the case of two Gaussians is given in closed form. More specifically, the KL-divergence between two uni-variate Gaussians $Q\sim\mathcal{N}\left(\mu_q,\sigma_q^2\right)$ and $P\sim\mathcal{N}\left(\mu_p,\sigma_p^2\right)$ is given by
\begin{equation}\label{eq:kl_div_gauss}
    D_{KL}\left(Q\Vert P\right) = \log\left(\frac{\sigma_q}{\sigma_p}\right) + \frac{\sigma_p^2+\left(\mu_q-\mu_p\right)^2}{2\sigma_q^2}-\frac{1}{2}.
\end{equation}

\begin{figure}[t]
\centering
    \includegraphics[width=0.65\linewidth]{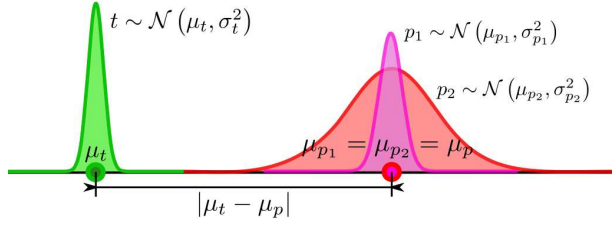}
    \caption{Comparison of KL divergence and $\mathcal{L}_1$ distance in measuring dissimilarity between distributions.}
    \label{fig:kl_div_intuition}
\end{figure}

In Fig.~\ref{fig:kl_div_intuition} we illustrate the behaviour of the KL divergence for measuring the dissimilarity between two pairs of uni-variate Gaussian distributions, namely $t\sim\mathcal{N}(\mu_t,\sigma_t^2)$ and $p_{1,2}\sim\mathcal{N}(\mu_{p_{1,2}},\sigma_{p_{1,2}}^2)$, where we assume that $\mu_{p_1}=\mu_{p_2}=\mu_p$. The KL divergence between $t$ and $p_1$ is larger than that between $t$ and $p_2$ (since $\sigma_2>\sigma_1$), indicating that $p_2$ is ``closer'' to $t$ than $p_1$. By contrast, a  distance metric which is typically employed as a regression loss function in localization tasks, such as the $\mathcal{L}_1$, when applied only on the mean values would assign the same distance from $t$ to $p_1$ and $p_2$, and thus the same regression loss.

Formally, we define our KL-based loss function $d_{kl}\colon\left(\mathbb{R}\times\mathbb{R}_+\right)\times\left(\mathbb{R}\times\mathbb{R}_+\right)\to\mathbb{R}_+$  between the distribution of the ground truth $\mathcal{N}\left(\mu_t,\sigma_t^2\right)$ and the distribution of the predictions $\mathcal{N}\left(\mu_p,\sigma_p^2\right)$ as
\begin{equation}\label{eq:kl_div_loss}
    d_{kl}\left((\mu_t,\sigma_t),(\mu_p,\sigma_p)\right) = \sqrt{\log\left(\frac{\sigma_p}{\sigma_t}\right) + \frac{\sigma_t^2+\left(\mu_q-\mu_p\right)^2}{2\sigma_p^2}-\frac{1}{2}}
\end{equation}

\begin{figure}[t]
\centering
    \includegraphics[width=0.6\linewidth]{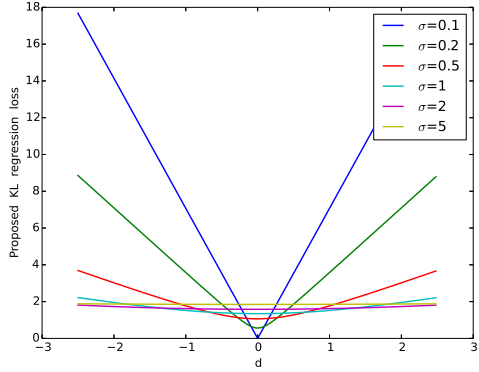}
    \caption{Proposed KL regression loss between given ground truth distribution $t\sim\mathcal{N}(\mu_t,\sigma_t^2)$ and a prediction distribution $p\sim\mathcal{N}(\mu_p,\sigma_p^2)$, where $\sigma_p^2\geq\sigma_t^2=0.01$. 
    KL loss degenerates to standard $\mathcal{L}_1$ loss for $\sigma_p^2=\sigma_t^2$.}
    \label{fig:kl_div_loss}
\end{figure}

In Fig.~\ref{fig:kl_div_loss} we plot the proposed KL loss function (\ref{eq:kl_div_loss}) between a given ground truth distribution $t\sim\mathcal{N}(\mu_t,\sigma_t^2)$, where $\sigma_p^2\geq\sigma_t^2=0.01$, and a prediction distribution $p\sim\mathcal{N}(\mu_p,\sigma_p^2)$ for different values of  $\mu_p$ and $\sigma_p^2$. We note that when $\sigma_p=\sigma_t$ KL loss admits the following form:
$$
d_{kl}\left((\mu_t,\sigma_t),(\mu_p,\sigma_t)\right) = \frac{\lvert\mu_t-\mu_p\rvert}{\sqrt{2\sigma_t^2}},
$$
which is a scaled variant of the standard $\mathcal{L}_1$ loss. In our problem, the proposed KL loss degenerates to $\mathcal{L}_1$ loss when prediction variances are equal or less than the variances of ground truth.

\section{Experiments}\label{sec:experiments}

In this section, we present the experimental evaluation of the proposed \van variants, as well as comparisons with state-of-the-art two-stage methods. First, we introduce the popular dataset that we use and the corresponding evaluation metrics that we adopt. Then, we evaluate our proposed Variance-Aware Network in three experimental settings (\vani, \vano, and \vanp) and we conduct an ablation study to investigate the effectiveness of different variants of our method. Note that, for \vanp, we only propagate the variances during training, while during testing only the means are propagated (similarly to the baseline). To gain more intuition about the second order modeling we use, we analyse the estimated/predicted variances. Finally, we compare our method with the state-of-the-art methods.

\subsection{Dataset and Evaluation}

\paragraph{\textbf{Dataset}} We evaluate the proposed method on the popular THUMOS'14~\cite{THUMOS14} dataset, which contains 200 and 213 temporal annotated untrimmed videos with 20 action classes in validation and testing set, respectively. Since there is no training dataset for it (UCF101~\cite{UCF101} is used instead), following the standard practice~\cite{zhao2017temporal,gao2017cascaded,chao2018rethinking}, we train our models on the validation set and evaluate them on the testing set.

\paragraph{\textbf{Evaluation metrics}} Similarly to several works in temporal action localization~\cite{shou2016temporal,shou2017cdc,gao2017cascaded,chao2018rethinking}, we report the mean Average Precision (mAP), where Average Precision (AP) is calculated for each action class. In addition, we report mAP at various temporal Intersection over Union (tIoU) thresholds (\ie, at tIoU in $\{0.3,0.4,0.5,0.6,0.7\}$) and the mean over them.

\subsection{Implementation Details}

Our baseline method is described in Sect.~\ref{subsec:baseline_method} and is briefly illustrated in Fig.~\ref{fig:overview}. The number of dimensions $d_i$ of the feature that feed the first FC layer after the pooling is $(k+2) \cdot 4096$ (+2 is used to utilize context feature before and after the given proposal).
The output of the first FC layer is $1000$ and it feeds the second FC layer, which extracts classification and regression scores as a $(C+1)\times3$ matrix, where $C=20$ is number of classes. During training, we use a batch size of $128$, and learning rate of $10^{-3}$. In all the reported results, we denote this standard setup as ``Baseline'', and we report the median of three experiments. We note that our baseline network degenerates to the network proposed in~\cite{xie2019exploring}. We train the network for $50K$ iterations.

\begin{table*}[t]
\centering
\small
\caption{Results (mAP@tIoUs (\%)) comparison among baseline and $\text{VAN}$s in temporal action localization on THUMOS'14.}
\label{tab:th14_map@tious}
\begin{tabular}{c|c|ccccc|c|c} \hline
    \multicolumn{1}{c}{k} & \multicolumn{1}{c}{tIoU} & \multicolumn{1}{l}{0.3} & \multicolumn{1}{l}{0.4} & \multicolumn{1}{l}{0.5} & \multicolumn{1}{l}{0.6}  & \multicolumn{1}{l}{0.7} &  \multicolumn{1}{l}{avg} & gaps \\ \hline
    \multirow{4}{*}{$1$} & Baseline &         51.03  &         43.41  &         33.16  &         20.72  &         10.48  &         31.76  & +0.00 \\ 
                         & \vani    &         50.17  &         42.98  &         33.37  &         20.92  &         10.25  &         31.54  & -0.22 \\ 
                         & \vano    &         51.67  &         44.02  &         33.72  & \textbf{21.83} &         11.28  &         32.50  & +0.74 \\
                         & \vanp    & \textbf{52.21} & \textbf{45.30} & \textbf{33.88} &         21.69  & \textbf{11.43} & \textbf{32.90} & +1.14 \\ \hline
    \multirow{4}{*}{$3$}  & Baseline & 54.88 & 47.21 & 37.22 & 25.04 & 12.70 & 35.41  & +0.00\\ 
    & \vani &  54.66 & 48.25 & 37.87 & 26.01 & 13.80 & 36.12 & +0.71 \\ 
    & \vano &  54.81 & 48.77 & 38.75 & 25.36 & 13.93 & 36.32 & +0.91\\
    & \vanp &  \textbf{55.66} &  \textbf{48.78} &  \textbf{38.81} &  \textbf{26.93} &  \textbf{15.21} &  \textbf{37.08} & +1.67 \\ \hline  
    \multirow{4}{*}{$4$}  & Baseline & 54.72 & 47.47 & 37.94 & 25.88 & 13.28 &	35.86 & +0.00 \\ 
    & \vani & 54.79 & \textbf{47.80} & 38.73 & 26.19 & 12.91 & 36.08 & +0.22\\ 
    & \vano & 54.17 & 47.72 & 38.24 & 24.20 & 12.94 & 35.45 & -0.41\\
    & \vanp & \textbf{54.92} & 47.75 & \textbf{38.98} & \textbf{26.99} & \textbf{14.29} & \textbf{36.59} & +0.73 \\ \hline
    \multirow{4}{*}{$5$}  & Baseline & 54.79 & 47.28 & 38.29 & 25.62 & 13.25 & 35.85 & +0.00 \\ 
    & \vani & \textbf{55.42} & \textbf{49.18} & 39.05 & 26.41 & 13.25 & 36.66 & +0.81\\ 
    & \vano & 55.11 & 47.96 & 39.03 & 25.62 & 14.68 & 36.48 & +0.63\\
    & \vanp & 55.03 & 48.56 & \textbf{39.17} & \textbf{26.90} & \textbf{14.97} & \textbf{36.93} & +1.08 \\ \hline
\end{tabular}
\end{table*}


\begin{table}[t]
    \begin{center}
    \caption{Number of parameters in different variations of VAN (k=3).} 
    \label{tab:num_params}
    \begin{tabular}{c|cccc}
      \hline
      Method       & Baseline & \vani   & \vano   & \vanp            \\ \hline
      \#parameters & 10.303M  & 20.543M & 10.366M & \textbf{10.303M} \\ \hline
    \end{tabular}
    \end{center}
\end{table}

\subsection{Experimental results}
\label{subsec:results}

In Table~\ref{tab:th14_map@tious}, we compare two different variants of the proposed method, namely \vano, and \vanp, in terms of mAP at various tIoU thresholds (\ie, $0.3,0.4,0.5,0.6,0.7$), for $k=1,3,4,5$. Also, to verify the effectiveness of \vanp in using input variances (\ie, whether it is meaningful to propagate input variances, instead of using them as extra features), we also report results of \vani. From the obtained results, we note that incorporating variance in general improves results, especially at tIoU thresholds of 0.5 or higher; that is, for more accurate localization. \vanp is almost always better than the baseline method by more than $1\%$. This indicates that the proposed \vanp can improve the performance of temporal action localization, and can effectively model the uncertainty by down-weighting the noisy training samples, so as to and enhance the robustness of the network.

\paragraph{\textbf{\vani vs \vanp}} Both networks use input variance (\ie, computed by the pooling operation) as an extra source of information in the input. While \vani use input variance as additional features, building a richer feature representation scheme (see Fig.~\ref{fig:proposed_method_intro}c), \vanp propagates it throughout the network so as to deliver probabilistic predictions in the output in the form of the mean and the variance  (see Fig.~\ref{fig:proposed_method_intro}d). Both of them improve the baseline. More specifically, \vanp is consistently better than \vani at avg(map@tIoU=0.3:0.7) and the threshold of 0.5, which is typically the threshold used for evaluating temporal action localization systems~\cite{shou2016temporal,shou2017cdc,yang2018exploring,gao2017cascaded,gao2017turn,chao2018rethinking}, and most often the case for higher thresholds as well. It is worth noting that \vani requires almost as many as double the number of parameters of \vanp, while the latter does not increase the number of parameters of the baseline.

\paragraph{\textbf{\vano vs \vanp}} Both methods use a KL-divergence-based loss function and model uncertainty in different ways. \vanp calculates uncertainty in terms of variance from the pooling operation, while \vano attempts to learn variances at the output (localization predictions). Table~\ref{tab:th14_map@tious} shows that \vanp outperforms \vano in general by taking extra information from the input. With respect to number of training parameters, \vano requires slightly more parameters than \vanp, due to the additional parameters introduced for learning output variances.

\begin{figure*}[!t]
    \centering
    \begin{subfigure}[b]{0.9\linewidth}
        \includegraphics[width=\textwidth]{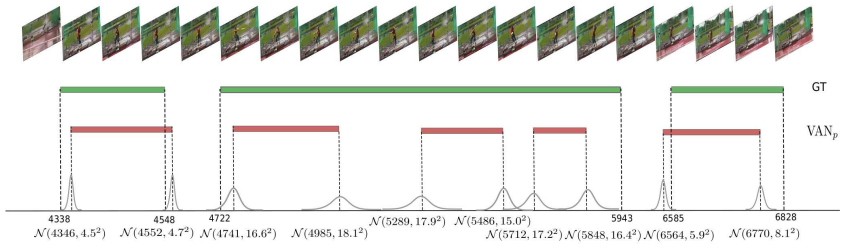}
        \caption{Part predictions of video\_test\_000372 (HammerThrow).}
        \label{subfig:video372}
    \end{subfigure}
    ~
    \begin{subfigure}[b]{0.85\linewidth}
        \includegraphics[width=\textwidth]{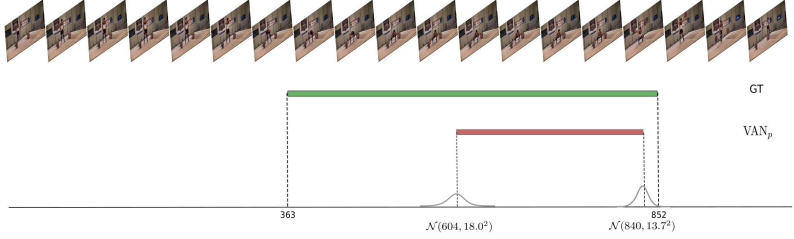}
        \caption{Part predictions of video\_test\_0001307 (CleanAndJerk).}
        \label{subfig:video1307}
    \end{subfigure}
     ~
    \begin{subfigure}[b]{0.85\linewidth}
        \includegraphics[width=\textwidth]{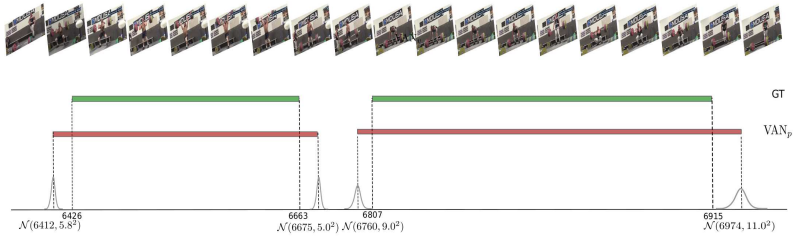}
        \caption{Part predictions of video\_test\_0001270 (CleanAndJerk).}
        \label{subfig:video1270}
    \end{subfigure}
    ~
    \begin{subfigure}[b]{0.85\linewidth}
        \includegraphics[width=\textwidth]{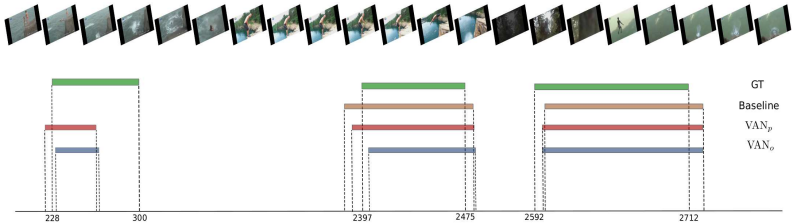}
        \caption{Comparison of different variation of the methods on video\_test\_0000357 (CliffDiving).}
        \label{subfig:video357}
    \end{subfigure}
\caption{Visualization of experimental testing results of different variations of \van: \textit{GT} (green bars) represents Ground Truth; \textit{Baseline} (brown bars) are the predictions from baseline network (k=3); \vanp (red bars) and \vano (blue bars) are the predictions from the corresponding network. In a),b),c), we put the predicted Gaussians there; Better viewed in color.}
\label{fig:video_examples}
\end{figure*}

Fig.~\ref{fig:video_examples} presents the predicted boundary distribution of \vanp in terms of means and variances of the start and the end of an action. Fig.~\ref{subfig:video372} shows that for proposals that are close to ground truth (\ie, the first and fifth in this case), the predicted variances of the boundaries are relatively low, while for the rest, inaccurate predictions, the predicted variances are larger, which indicates the attempt of the network to cover the whole action. This is also shown in the following examples. Fig.~\ref{subfig:video1307} shows a complete \textit{CleanAndJerk} action with a long preparation, while Fig.~\ref{subfig:video1270} (the left ground truth action) shows a complete one with normal speed. This shows that there is a large variation in the way that actions are performed -- in the case of the long preparation that there is not much motion in the case of the preparation phase of the \textit{CleanAndJerk} action, this results to the misdetection of the start of the action and higher uncertainty (large variances) in the corresponding estimation. 

Fig.~\ref{subfig:video1270} depicts two instances of \textit{CleanAndJerk}, where in the second one the athlete fails to perform the jerk properly and our model tends to predict larger uncertainty for the end. Finally, Fig.~\ref{subfig:video357} shows the predictions of \vanp, \vano, and the baseline method, indicating that modeling boundary uncertainty using uni-variate Gaussians is beneficial to localization accuracy by capturing and exploiting inherent uncertainty of the data and/or the model.

\begin{figure*}[t]
    \centering
    \begin{subfigure}[b]{0.7\linewidth}
        \includegraphics[width=\textwidth]{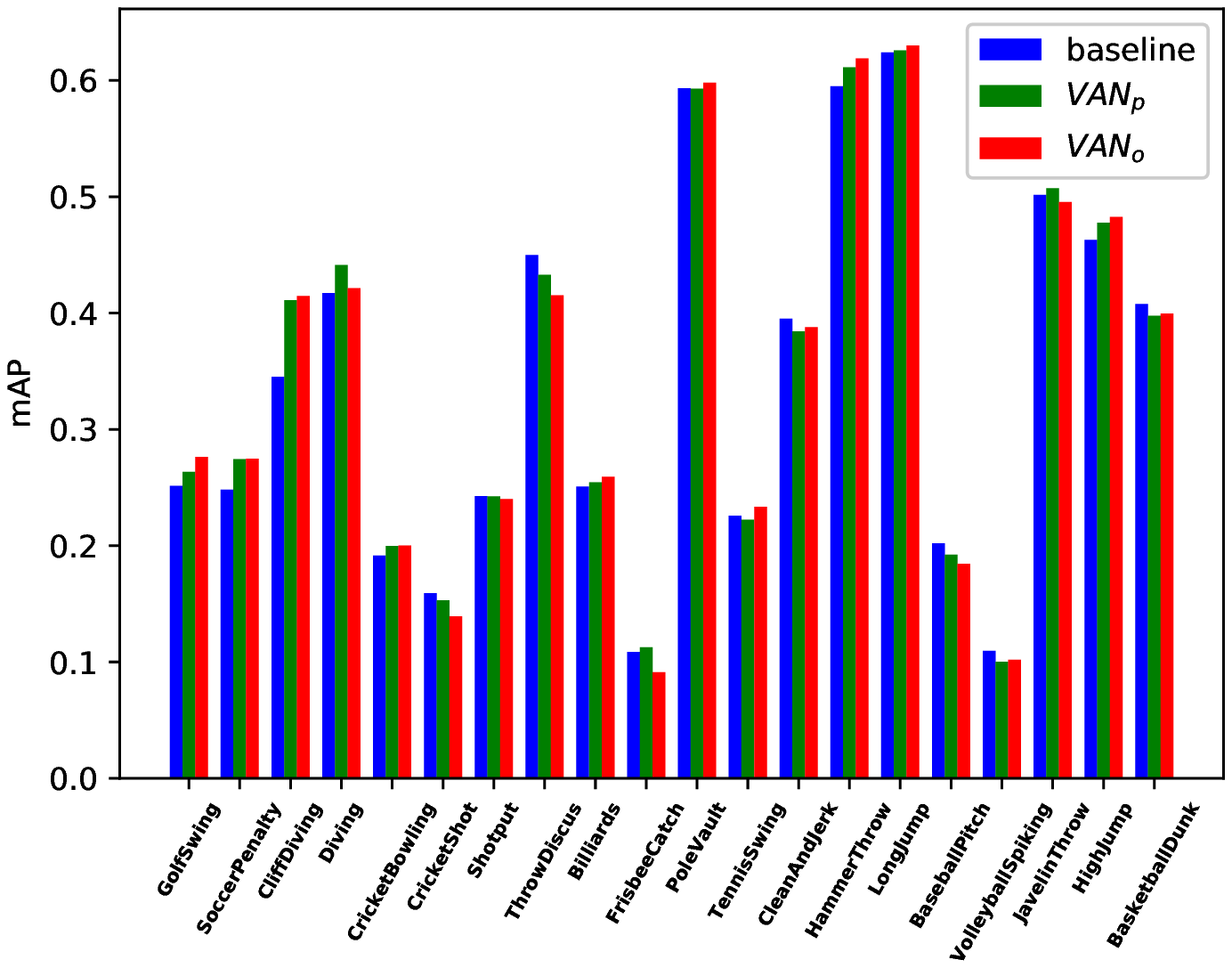}
        \caption{mean Average Precision(mAP) comparison for each class.}
        \label{subfig:map_comp}
    \end{subfigure}
    \begin{subfigure}[b]{0.48\linewidth}
        \includegraphics[width=\textwidth]{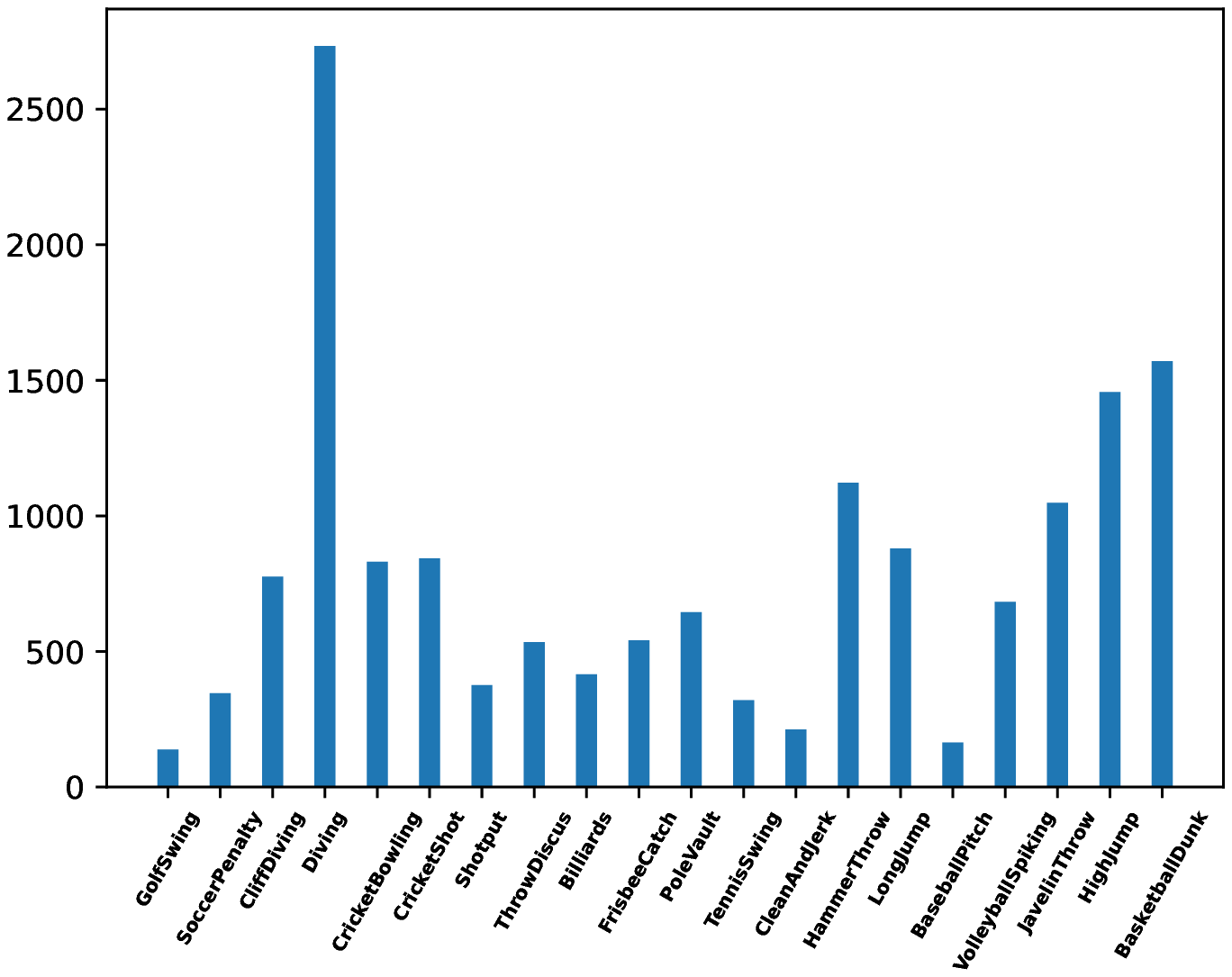}
        \caption{Number of training samples for each class.}
        \label{subfig:number_sample}
    \end{subfigure} 
    ~
    \begin{subfigure}[b]{0.48\linewidth}
        \includegraphics[width=\textwidth]{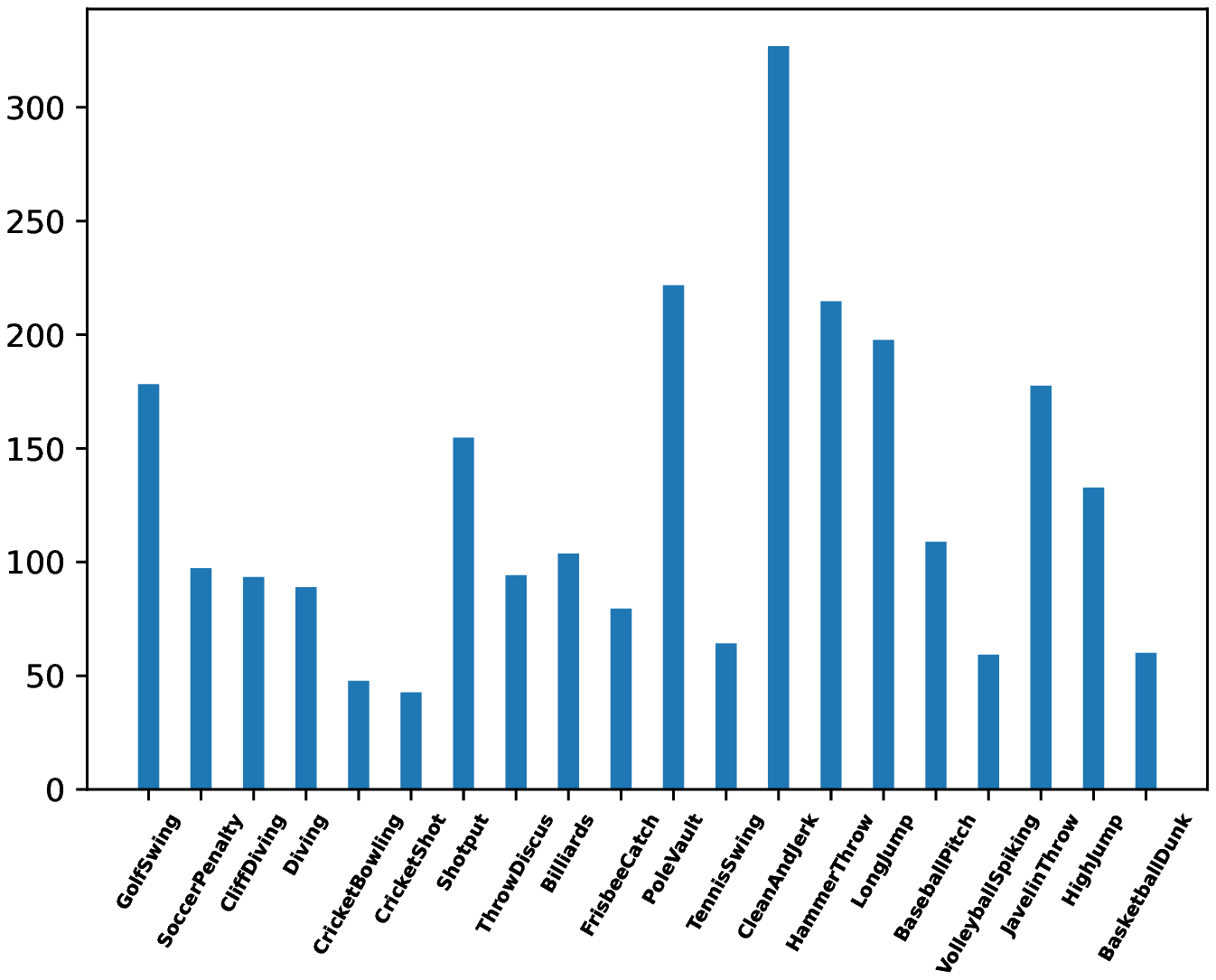}
        \caption{Length of actions in each class.}
        \label{subfig:length_sample}
    \end{subfigure}
\caption{Stastics figures. a) Methods performance comparison for different action category in THUMOS14; b) Number of training samples comparison; c) Average length of training samples comparison. Better viewed in color. }
\label{fig:stastics}
\end{figure*}
We report the mAP (mean Average Precision) regarding to each class in THUMOS'14 in Fig.~\ref{subfig:map_comp}, and also number of training samples distribution/length distribution in Fig.~\ref{subfig:number_sample} and \ref{subfig:length_sample}. It clearly shows that compared with baseline, both \vanp and \vano consistently improve the performance by loss attenuation, especially for long actions with enough training samples, such as \textit{Diving}, \textit{CliffDiving}, \textit{HammerThrow}, \textit{HighJump}, \textit{GolfSwing}, etc. For short actions, especially when the number of available training samples is limited, such as \textit{CricketShot} and \textit{BaseballPitch}, the proposed methods does not apply well by down-weighting the loss of these corresponding training samples. Finally, we note that \vanp outperforms \vano in general.

\subsection{State-of-the-art comparisons} 

Finally, in Table~\ref{tab:exp_results_soa}, we report the experimental results of \vanp compared to other state-of-the-art works. To make a fair comparison, we did not include methods using Graph Neural Network based detectors~\cite{xu2020g,zeng2019graph}. We note that the proposed method, even though it is simple, outperforms other state-of-the-art TAL methods, especially those works~\cite{gao2017turn,gao2017cascaded,xie2019exploring} that use the same proposal generator (TURN)~\cite{gao2017turn} with us by a large margin. 

\begin{table}[t]
    \small
    \caption{State-of-the-art comparison on Temporal action localization on THUMOS'14 for various tIoU thresholds.}
    \label{tab:exp_results_soa}
	\begin{center}
	    \begin{tabular}{lccccccc} \hline
        tIoU & 0.3  & 0.4  & 0.5  & 0.6  & 0.7  \\ \hline
        \multicolumn{1}{l}{SCNN~\cite{shou2016temporal}}     & 36.3 & 28.7 & 19.0 & 10.3 & 5.3  \\ 
        \multicolumn{1}{l}{Glimpse~\cite{yeung2016end}}        & 36.0 & 26.4 & 17.1 & --   & --   \\ 
        \multicolumn{1}{l}{CDC~\cite{shou2017cdc}}          & 40.1 & 29.4 & 23.3 & 13.1 & 7.9    \\ 
        \multicolumn{1}{l}{SNN~\cite{zhao2017temporal}}     & 51.9 & 41.0 & 29.8 & --   & --   \\ 
        \multicolumn{1}{l}{BSN~\cite{lin2018bsn}}  & 53.5 & 45.0 & 36.9 & 28.4 & 20.0  \\ 
        \multicolumn{1}{l}{TAL-Net~\cite{chao2018rethinking}}   & 53.2 & 48.5 & 42.8 & 33.8 & 20.8   \\ 
        \multicolumn{1}{l}{BMN~\cite{lin2019bmn}} & 56.0 & 47.4 & 38.8 & 29.7 & 20.5\\ 
        \multicolumn{1}{l}{GTAN~\cite{long2019gaussian}} & 57.8 & 47.2 & 38.8 & - & - \\ \hline 
        \multicolumn{6}{c}{\textit{Same proposals}} \\ \hline
        TURN~\cite{gao2017turn} &  44.1 & 34.9 & 25.6 & - & - \\ 
        Cascade~\cite{gao2017cascaded}  & 50.1 & 41.3 & 31.0 & 19.1 & 9.9    \\ 
        TAD~\cite{xie2019exploring} & 51.7 & 46.6 & 36.8 & 25.4 & 12.7 \\ 
        \vano ($k=5$) & \textbf{55.1} & 48.0 & 39.0 & 25.6 & 14.7 \\ 
        \vanp ($k=5$) & 55.0 & \textbf{48.6} & \textbf{39.2} & \textbf{26.9} & \textbf{15.0}  \\ \hline 
        \end{tabular}
    \end{center}
\end{table}

\section{Conclusions}\label{sec:conclusions}

In this paper, we proposed a Variance-Aware Network (VAN) for the problem of temporal action localization that take into consideration second-order statistics, either in the input or in the output of regression tasks. In this way, information that would be lost during the pooling operation is preserved and used for improving and robustifying localization. For doing so, we derived analytical or reasonable approximations on how input uncertainty (in terms of feature variance) is propagated throughout the network, leading to probabilistic predictions (action boundaries). Moreover, we designed a KL-divergence-based loss function that allowed for end-to-end training using standard back-propagation. We showed that the proposed method lead to large improvements in comparison to the baseline and other state-of-the-art two-stage approaches.


%
%

\bibliographystyle{spmpsci}      
\bibliography{Manuscript}


\end{document}